\documentclass{sig-alternate-05-2015}

\usepackage{epsfig}
\usepackage{epstopdf}
\usepackage{graphicx}
\usepackage{amsmath,amssymb} 
\usepackage{array}
\usepackage{algorithm}
\usepackage{algorithmic}

\begin{document}

\setcopyright{acmcopyright}

\CopyrightYear{2016} 
\setcopyright{licensedusgovmixed}
\conferenceinfo{UrbanGIS 16,}{October 31-November 03 2016, Burlingame, CA, USA}
\isbn{978-1-4503-4583-5/16/10}\acmPrice{\$15.00}
\doi{http://dx.doi.org/10.1145/3007540.3007548}

\doi{10.475/123_4}

\isbn{123-4567-24-567/08/06}


\acmPrice{\$15.00}

%

\title{Combining Maps and Street Level Images for Building Height and Facade Estimation
\titlenote{This manuscript has been authored by UT-Battelle, LLC under Contract No. DE-AC05-00OR22725 with the U.S. Department of Energy.  The United States Government retains and the publisher, by accepting the article for publication, acknowledges that the United States Government retains a non-exclusive, paid-up, irrevocable, world-wide license to publish or reproduce the published form of this manuscript, or allow others to do so, for United States Government purposes.}}
%
%
%
%
%

\numberofauthors{2} 
%
\author{
\alignauthor
Jiangye Yuan\\
       \affaddr{Computational Sciences \& Engineering Division}\\
       \affaddr{Oak Ridge National Laboratory}\\
       \affaddr{Oak Ridge, Tennessee 37831}\\
       \email{yuanj@ornl.gov}
\alignauthor
Anil M. Cheriyadat\\
       \affaddr{Computational Sciences \& Engineering Division}\\
       \affaddr{Oak Ridge National Laboratory}\\
       \affaddr{Oak Ridge, Tennessee 37831}\\
       \email{cheriyadatam@ornl.gov}
}

\maketitle
\begin{abstract}
We propose a method that integrates two widely available data sources, building footprints from 2D maps and street level images, to derive valuable information that is generally difficult to acquire -- building heights and building facade masks in images. Building footprints are elevated in world coordinates and projected onto images. Building heights are estimated by scoring projected footprints based on their alignment with building features in images. Building footprints with estimated heights can be converted to simple 3D building models, which are projected back to images to identify buildings. In this procedure, accurate camera projections are critical. However, camera position errors inherited from external sensors commonly exist, which adversely affect results. We derive a solution to precisely locate cameras on maps using correspondence between image features and building footprints. Experiments on real-world datasets show the promise of our method. 
\end{abstract}

%
%

%
%

%
%
\printccsdesc



\section{Introduction}

\begin{figure}
\begin{center}
\includegraphics[width=0.48\textwidth]{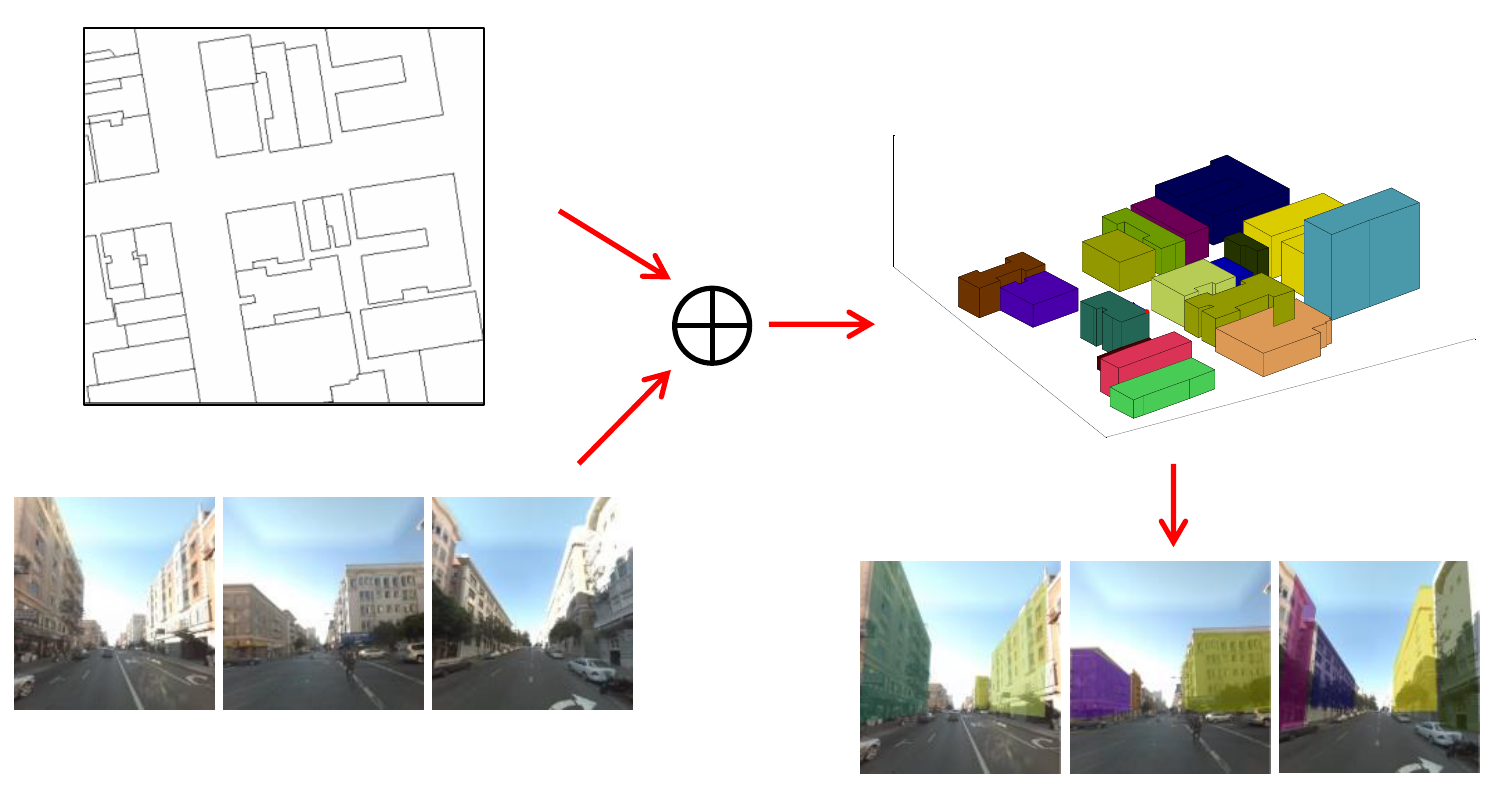}
\end{center}
 \caption{Building height estimation and building facade identification using building footprints and street level images.}
\label{fig:overview}
\end{figure}

The rapid development of data collection capabilities gives rise to multi-modal datasets. Fusing the available data leads to new solutions to problems that are challenging, even impossible, with single-modal datasets. In particular, integrating geographic information with image data has been increasingly exploited \cite{yuan2013,wang2013,wang2015}, which significantly enhances the information extraction capability. In this paper, we utilize building footprint data from GIS maps and street level images, and develop a fully automatic method to estimate building heights and generate building facade masks in images. The concept is illustrated in Fig~\ref{fig:overview}. 

Building heights are important information for many applications related to urban modeling. Stereo pairs of high resolution aerial images and aerial LiDAR data are typically used to acquire height information \cite{zebedin2008,tack2012,chen2008}. Although highly accurate results can be obtained, those types of data are not easily accessible due to expensive collection process. We derive height information in a novel way. Through camera projections, we map edges of building footprints with different heights to images. The building height is determined by matching the projected edges with building rooflines in street view images. We design an indicator that incorporates color and texture information and reliably find the projected edges aligned with rooflines. This method does not require aerial sensing data. In addition, by extending building footprints vertically with estimated heights, we can generate 3D building models. Despite being significantly simplified, such building models are useful in applications requiring low storage and fast rendering.

Identifying buildings in images is an important yet challenging task for scene understanding. A main strategy is to learn a classifier from labeled data \cite{xiao2009,tighe2010}. Deep neural networks trained with massive labeled data have shown excellent performance on segmenting semantic objects in images, including buildings \cite{farabet2013,long2015}. In contrast, we generate building masks by projecting the 3D building models onto images. This approach provides an alternative solution that has the following advantages. First, it is capable of dealing with large appearance variations without complicated training and expensive labeled data collection. Second, each individual building in images is associated with a building on maps. Such a link allows attribute information from map data to be easily transfered to the images. For example, we can identify the buildings in an image corresponding to a specific restaurant if the information is given on maps. Note that while extracting builds from images has been addressed through a similar strategy of projecting georeferenced models~\cite{cho2013,wang2013}, those studies use existing building models instead of 2D maps.  


Our approach involves matching building features between maps and images. However, even when camera projection parameters are available from sensors, cross alignment between maps and images still poses a major challenge. A main reason is that the measurement of camera positions mostly relies on Global Positioning Systems (GPS), which tend to be affected by measurement conditions, especially in high density urban areas. Despite various new techniques combined with additional sensory information \cite{qi2002,drawil2010,jo2012}, GPS errors remain at a noticeable level. For example, a median position error of 5-8.5 meters is reported for current generation smartphones \cite{zandbergen2011}. To overcome inaccurate spatial alignments, we propose an effective approach based on correspondence between map and image features to refine camera positions on maps. The method is fully automated and works reliably on real-world data.. 

In this work, we use building footprint layers from OpenStreetMap (OSM)\footnote{http://www.openstreetmap.org/} -- the largest crowd sourced maps produced by contributors using aerial imagery and low-tech field maps. The OSM platform has millions of contributors and is able to generate map data with high efficiency. We use Google Street View images, which capture street scenes with a world-wide coverage. Images can be downloaded through an publicly available API\footnote{https://developers.google.com/maps/documentation/ streetview/} with all camera parameters provided.  

\section{Related work}

A number of previous studies have explored the idea of creating building models based on 2D maps, where a critical step is to obtain height information. In \cite{haala1996}, building heights and roof shapes are set to a few fixed values based on known building usage. In more recent work, stereo pairs of high resolution aerial images and aerial LiDAR data are used to estimate height information \cite{zebedin2008,tack2012, chen2008}. However, since those types of data are not available for most areas in the world, the methods do not scale well. In this work, a new method is proposed to acquire building heights, which utilizes map data and street view images. Since input data are widely available, the method can scale up to very large areas.  

In order to correctly project map data onto images, images need to be registered with maps. This is a difficult task  because it requires matching between abstract shapes on a ground plane and images from very different views. This has been pursued in a few recent studies. In \cite{cham2010} omnidirectional images are used to extract building corner edges and plane normals of neighboring walls, which are then compared with structures on maps to identify camera positions. The method in \cite{chu2014} follows the same framework and aims to refine camera positions initially from GPS. It works on a single view image, but it involves manual segmentation of buildings in images to obtain highly accurate building corner lines. Instead of 2D maps, digital elevation models (DEM) has also been utilized, where building roof corners are extracted and matched with those in images \cite{bansal2014}. However, DEM is much less accessible than maps. It should be noted that existing techniques for camera pose estimation given 2D-to-3D point correspondences (the P$n$P problem) are not applicable here because point correspondences are not available. We propose a method that registers a single view image with maps through a voting scheme. The method is fully automated and works reliably on real-world data. 

The rest of the paper is organized as follows. Section~\ref{sec:bldmdl} presents the method to estimate building heights and identify buildings in images. The method for estimating accurate camera position on maps is discussed in Section~\ref{sec:locenh}. In Section~\ref{sec:Experiments} we conduct experiments on large datasets and provide quantitative evaluation. We conclude in Section~\ref{sec:Conclusions}.

\section{Building height and facade estimation}
\label{sec:bldmdl}

\begin{figure}
\begin{center}
\includegraphics[width=0.21\textwidth]{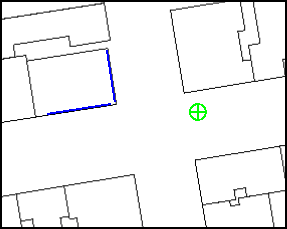}
\hspace{2mm}
\includegraphics[trim = 0mm 60mm 0mm 20mm, clip, width=0.21\textwidth]{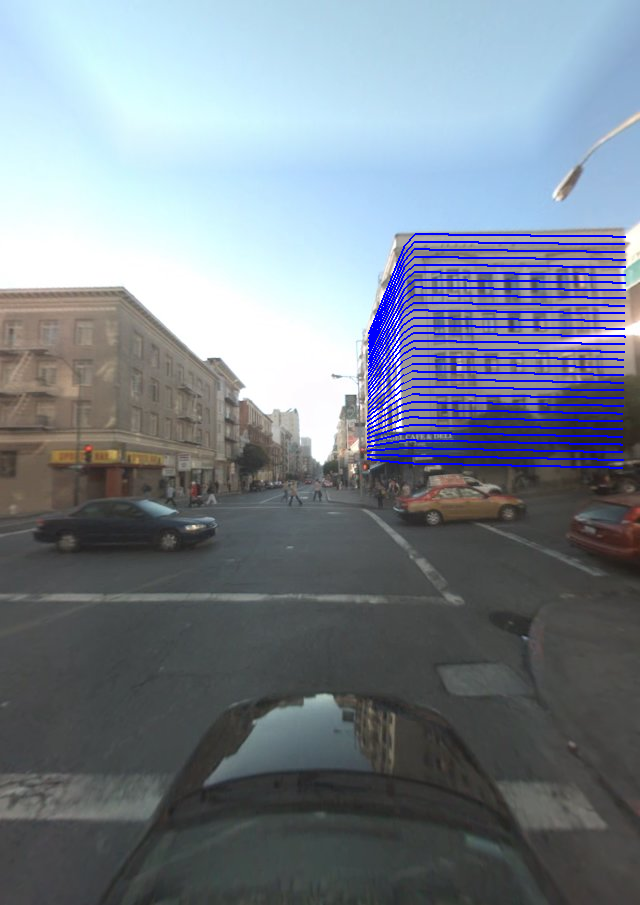}

\makebox[0.21\textwidth][c]{(a)}
\makebox[0.21\textwidth][c]{(b)}
\end{center}
 \caption{Building height estimation. (a) A map containing building footprints. The green marker represents the camera location. Blue edges indicate the part within the field of view. (b) A street level image. All blue lines are the projected edges with different elevation values. }
\label{fig:bldht}
\end{figure}

For a building footprint in a georeferenced map, we have access to 2D geographic coordinates (e.g., latitude and longitude) of the building footprint. If we set the elevation of the building footprint to the ground elevation, we have 3D world coordinates. Then we project the edges of the building footprint onto the image through camera projections. The projected edges should outline the building extent at its bottom. As we increase the elevation of the building footprint, the projected edges move toward the building roof. When the projected edges are aligned with building rooflines, the corresponding elevation is the roof elevation of the building. The building height is simply the increased elevation. This procedure is illustrated in Fig.~\ref{fig:bldht}, where we project footprint edges within the field of view.   

We need to determine whether projected footprint edges are aligned with building rooflines. A straightforward way is to compute image gradients and select the projected edges with the maximum gradient magnitude as the roofline. However, because lighting conditions change significantly and there exist many straight edges other than rooflines, gradient magnitude often fails to represent the existence of rooflines.   

\begin{figure}
\begin{center}
\includegraphics[width=0.45\textwidth]{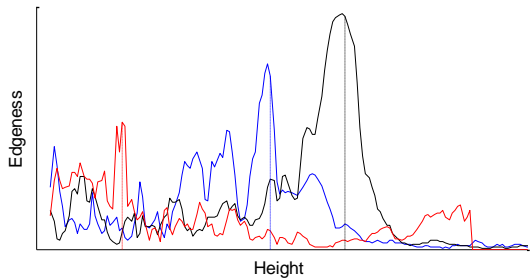}
\end{center}
 \caption{Height estimation based on edgeness scores. Each color represents one building. Dashed lines indicate actual heights of buildings.  }
\label{fig:edgn}
\end{figure}

Here we use an edgeness indicator that incorporates both color and texture information \cite{liu2002}. It is fast to compute and performs reliably. We first compute spectral histograms at a local window around each pixel location (the window size is set to $17\times17$ in our experiments). A spectral histogram of an image window is a feature vector consisting of histograms of different filters responses. It has been shown that spectral histograms are powerful to characterize image appearances \cite{liu2002}. Here we use a spectral histogram concatenating histograms of RGB bands and two Laplacian of Gaussian (LoG) filter responses. Two LoG filters are applied to the grayscale image converted from RGB bands ($\sigma$ is set to 0.5 and 1, respectively). The histogram of each band has 11 equally spaced bins. Local histograms can be computed efficiently using the integral histogram approach. Since histograms of RGB bands are sensitive to lighting conditions, we weight those histograms by 0.5. The edgeness indicator value at $(x, y)$ is then defined as the sum of two feature distances between pixel locations $<(x + h, y), (x - h, y)>$ and $<(x, y + h), (x, y - h)>$, where $h$ is half of the window side length. $\chi^2$ difference is used as the distance metric. This edgeness indicator reflects the appearance change between neighboring windows and thus better captures rooflines, which are essentially boundaries separating two distinctive regions. Fig.~\ref{fig:edgn} shows a plot of edgeness scores versus heights for three buildings, where actual building heights are well captured by maximum edgeness.  

The main steps of the proposed method are described as follows. 

\begin{enumerate}
\item \textit{Identify visible building edges on a map.} Based on camera parameters, we determine the field of view on the map. We select the edges of building footprints that are not occluded by other buildings. Buildings far away from the camera may not show clear rooflines in an image. Therefore, we only consider buildings within a distance of 60 meters.   

\item \textit{Project selected edges onto the image.} 2D geographic coordinates of the edges are known. We assign ground elevations to the edges and map them onto the image through a projective camera projection, which results in polylines in the image. Highly accurate ground elevations can be obtained from public geographic databases.

\item \textit{Determine building heights.} For each visible building, we gradually increase the elevation of selected edges and project them onto the image. The projected edges scan the image through a series of polylines. When the sum of the edgeness scores on a polyline reaches the maximum, the corresponding elevation gives the building height. There are cases where taller buildings behind the target building are also visible, but they usually have rooflines with different length and shape, which do not give maximum edgeness values.     

\item \textit{Process a set of images.} The height of a building can be estimated sufficiently well with one image as long as building rooflines are visible in the image. If a collection of images are available, where a building appears in multiple images, we can utilize the redundancy to improve accuracy. For each building, we create a one-dimensional array with each element representing a height value and initialize all elements to zeros. After scanning an image for a building, we add one to the element corresponding to the estimated height value. Once all images containing the building are processed, we choose the element with the maximum value to obtain the building height. In some images where only lower part of a building can be seen, estimated heights are incorrect. We use a simple technique to deal with this issue. Because in such a case the projected edges are within building facades, the edgeness scores are generally small. We ignore the estimated height if the corresponding indicator value is smaller than a threshold, and thus incorrect estimates are not recorded in the array.  

\item \textit{Generate building facade masks in images.} Simple 3D buildings models can be generated by extruding boxes from building footprints with estimated heights. Buildings in the field of view are projected back onto images, which results in labeling facades of each building. To address the visibility problem (building models partially occluded by others), we use the painter's algorithm, where the farthest facade is projected first.       
  
\end{enumerate}

Our method treats building roofs as flat planes. For buildings with other roof shapes, the resulting height is generally between the top and the base of roofs. Since most non-flat roofs have a small height, our method can still give an estimate close to the mean height. The method may not work well when building rooflines in an image are completely occluded by other objects (e.g., trees and other buildings). However, with multiple views available, the method can exploit the rooflines of the same building visible in other images taken from different angles. If a building is completely obstructed on the map but visible in images because it is taller than obstructing buildings, the method will skip the building.   

\section{Camera localization on maps}
\label{sec:locenh}

\begin{figure}
\begin{center}
\includegraphics[trim = 0mm 60mm 0mm 0mm, clip, width=0.15\textwidth]{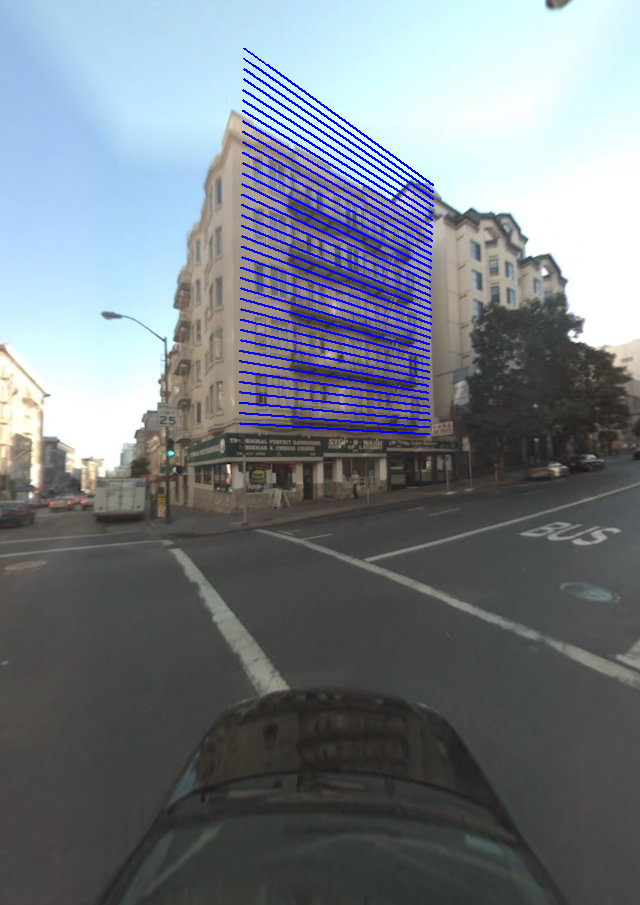}
\includegraphics[trim = 0mm 60mm 0mm 0mm, clip,width=0.15\textwidth]{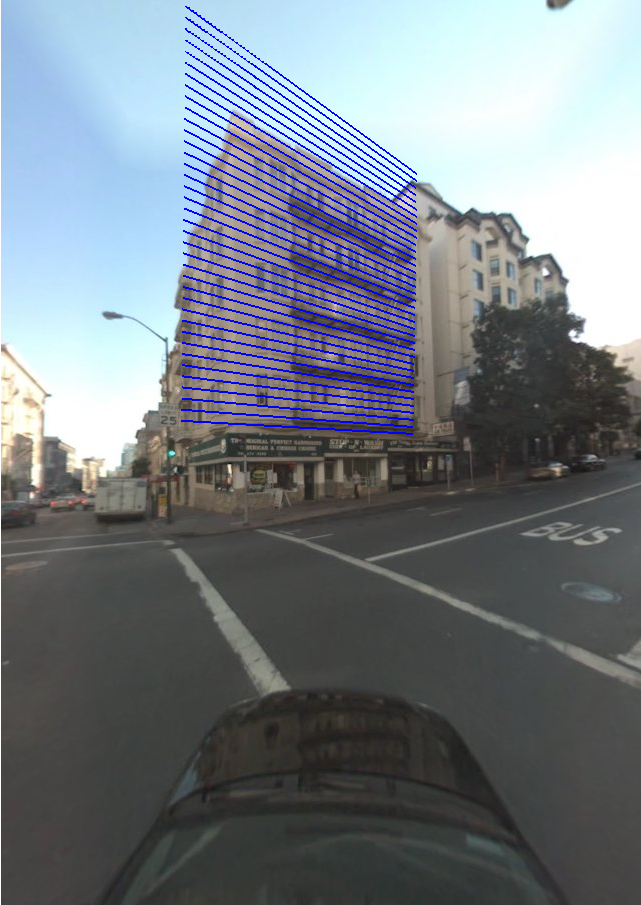}
\includegraphics[trim = 0mm 60mm 0mm 0mm, clip,width=0.15\textwidth]{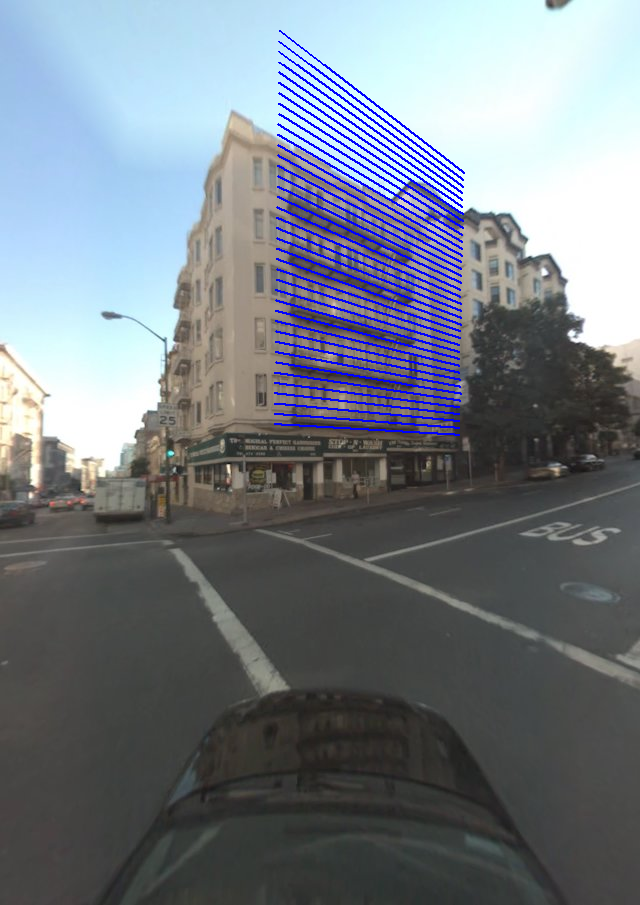}

\makebox[0.15\textwidth][c]{(a)}
\makebox[0.15\textwidth][c]{(b)}
\makebox[0.15\textwidth][c]{(c)}
\end{center}
 \caption{Impact of camera position errors. (a) A building footprint edge projected onto an image with a correct camera position. A series of elevation values are used. (b) and (c) The projections with camera positions moved 3 meters away from the correct one. }
\label{fig:locoff}
\end{figure}

As discussed earlier, we need to register images with maps for accurate projections. In this paper, we focus on camera positions, the measurement of which is much more noisy than orientation measurements. To illustrate the effect of camera position errors on projection results, Fig.~\ref{fig:locoff} shows the building footprint edges projected onto images with shifted camera positions. As can be seen, there are clear misalignments between projected footprint edges and buildings in images even though camera positions are shifted by only 3 meters. Such misalignments may cause projected building footprints to fail to match the corresponding rooflines and hence incorrect height estimation. They also lead to incorrect building facade masks. In the following we propose a method that takes an initial position that may be noisy and estimate the accurate position on maps.

\subsection{Camera position from point correspondence}

A 3D point $\mathbf{P}=(X,Y,Z)^T$ in world coordinates can be projected onto a pixel location $p=(u,v,1)^T$ in an image plane through the camera projection equation:
\begin{equation}
\lambda\mathbf{p}=\left[ \mathbf{K} | \mathbf{0}_3 \right]
\begin{bmatrix}
\mathbf{R} & - \mathbf{R}\mathbf{C}\\ 
\mathbf{0}_3^T & 1
\end{bmatrix}
\begin{bmatrix}
 \mathbf{P} \\
1
\end{bmatrix}
=\mathbf{K}\mathbf{R}\mathbf{P}- \mathbf{K}\mathbf{R}\mathbf{C}.
\label{cp}
\end{equation} 
$\lambda$ is the scaling factor, $\mathbf{K}$ the camera intrinsic matrix, $\mathbf{R}$ the camera rotation matrix, and $\mathbf{C}$ the location of the camera center in world coordinates. Note that $-\mathbf{R}\mathbf{C}$ equals to the camera translation. Given a pair of $p$ and $P$, we have 
\begin{equation}
\mathbf{C'}=\mathbf{P}+\lambda\mathbf{R}^{-1}\mathbf{K}^{-1}\mathbf{p}.
\label{cl}
\end{equation} 
That is, given correspondence between a pixel location and a 3D point in world coordinates, the possible camera positions $\mathbf{C'}$ lie on a 3D line defined by (\ref{cl}).    

In this work, we assume that camera position errors are mainly in a horizontal plane, because vertical camera positions can be reliably obtained with existing ground elevation data. Let $\mathbf{C'}=(X_C,Y_C,Z_C)^T$. We can discard Z dimension and simplify (\ref{cl}) to 
\begin{equation}
\begin{bmatrix}
X_C \\
Y_C
\end{bmatrix}
=
\begin{bmatrix}
X \\
Y
\end{bmatrix}
+\lambda
\begin{bmatrix}
\Delta X \\
\Delta Y
\end{bmatrix},
\end{equation} 
where $\Delta X$ and $\Delta Y$ are truncated from
$(\Delta X,\Delta Y, \Delta Z)^T=\mathbf{R}^{-1}\mathbf{K}^{-1}\mathbf{p}$. This defines a line on a 2D plane. If correspondence of two pairs of points is given, the camera position can be uniquely determined, which is the intersection of two lines. 

\subsection{Camera position from image and map correspondence}

Based on the above analysis, we can determine camera positions based on point correspondence between images and maps. We use corners on building footprints, which can be easily identified from map data. However, it is difficult to find the corresponding points in an image. To provide better features for matching, we place a vertical line segment on each building footprint corner (two end points have the same 2D coordinate but different elevation values). We will refer to such a line as a building corner line (BCL). The line length is fixed. Note that at this stage building height is not available. When projecting a BCL onto an image with an accurate camera position, it should be well aligned with building edges in the image. 

\begin{figure*}
\begin{center}
\includegraphics[width=0.95\textwidth]{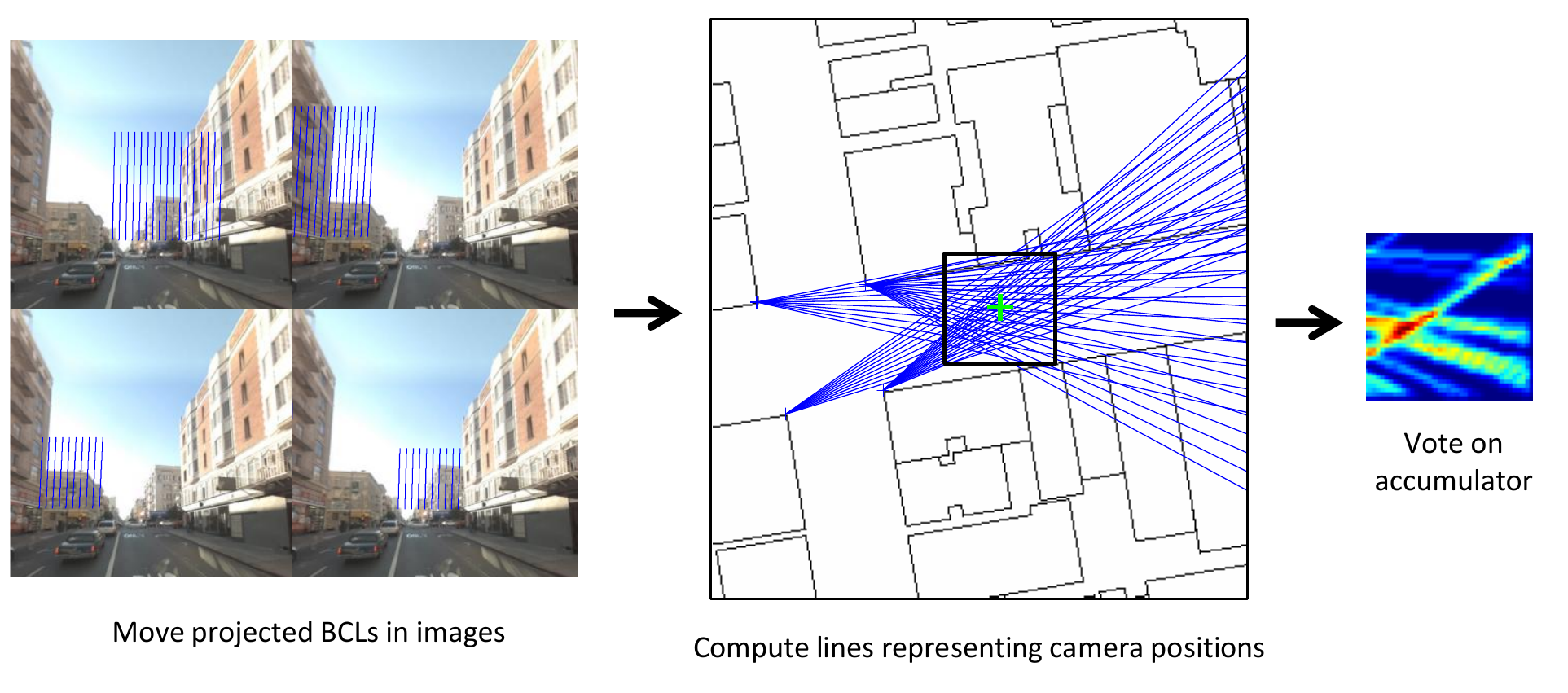}
\end{center}
 \caption{Illustration of camera position estimation. By moving projected BCLs (blues lines) in images, we compute potential camera positions (blue lines) on the map, which add values to an accumulator based on alignment measure between projected BCLs and building edges. The green marker indicates the initial camera position, and the black window shows the extend of the accumulator. In the accumulator, red pixels represent large values.  }
\label{fig:locest}
\end{figure*}

Under the assumption that position errors are within a horizontal plane, for a BCL projected onto an image with an inaccurate camera position, the displacement is along image columns. We first project a BCL onto an image using an initial camera position. Next, we horizontally move the projected BCL toward both directions within a certain range. The range is set to be inversely proportional to the distance from the BCL to the camera position. At each moving step, we have a pair of $\mathbf{P}$, the building footprint corner, and $\mathbf{p}$, the projected point moved along columns. By using (\ref{cl}), we can compute a line on the map that represents potential camera positions. We create an accumulator, which is a 2D array centered at the initial camera position. For each line, we increment the value of the bins the line passes through. The increased value is determined based on how well the moved BCL is aligned with building edges. At the end, the bin with the highest value gives the most likely camera position, which results in the best match between BCLs and building edges in images. The procedure is illustrated in Fig.~\ref{fig:locest}.

When moving a projected BCL, we need to know whether it is aligned with a building edge so that proper values can be assigned to the accumulator. Detecting building edge is another challenging task. In an image of a street scene, contrast of building edges varies significantly due to changes of lighting conditions. Building edges can be fully or partially occluded by other objects. Moreover, there exist a large number of edges from other objects. Although the voting-based method is able to tolerate a reasonable amount of errors in detection results, a large number of false positives and false negatives can still confuse the method. 

We exploit the idea of line support regions \cite{burns1986}, which is widely used for straight line extraction. A line support region consists of pixels with similar gradient orientations, where a line segment can be extracted. A common practice for building edge extraction in previous work \cite{cham2010,chu2014} is to assume that building edges have relatively large contrast. However, we observe that in many cases there are only slight color changes around building edges. Line support regions are formed based on consistency of gradient orientations regardless of magnitudes and hence better suited in this task.     

Since street level images are captured by a camera with its Y-axis orthogonal to the ground, building edges in images should be close to vertical lines. We select pixels with gradient orientations within $22.5^{\circ}$ around the vertical direction and find connected regions with large vertical extents and small horizontal extents (in the experiments two thresholds are set to 50 and 20 pixels, respectively). When a projected BCL hits a region center, the corresponding accumulator bins increment by one. A Gaussian density kernel is placed on each region center, and a projected BCL close to a center point casts a value equal to the maximum density value it reaches.

\subsection{Integration with height estimation}
Given an image and a map, the procedure includes the following two steps. The first is to apply the voting based method to determine the camera position, and the second is to estimate the building height by examining projected building footprints. In certain scenarios, a set of line support regions that do not correspond to building edges happen to form a strong peak, resulting in an incorrect camera location. To address this issue, we integrate height estimation with camera position estimation, which utilizes the information of building footprint edges in addition to corners to obtain accurate positions and simultaneously produces building heights.  

From a resulting accumulator, we select the top $k$ local maxima as candidate camera positions. For each of them, we project building footprint edges to match building rooflines in an image, as described in Section~\ref{sec:bldmdl}. We calculate the sum of edgeness scores for all projected building footprint edges that matches rooflines. We select the candidate with the largest sum, where rooflines in the image should be best aligned with building footprints. This strategy integrates two separate steps and reduces ambiguities in correspondence between BCLs and building edges. Algorithm \ref{alg1} summarizes the algorithm for processing a street view image. 

\begin{algorithm}                     
\caption{Building height estimation with camera position refinement}  
\begin{algorithmic}[1] 
 \STATE Extract line support regions
 \STATE Compute an accumulator using the voting method. Select the locations of top k peaks ${pos(i),i=1,\ldots,k}$
\FOR{$i$:=1 \TO $k$ }
  \STATE Set camera position to $pos(i)$
  \STATE Identify visible buildings on map
 \FOR{each visible building}  
   \STATE Project building footprints onto image and estimate height based on edgeness values
 \ENDFOR
 \STATE Compute $S(i)$: the sum of edgeness values of projected edges matching rooflines
\ENDFOR
 \STATE Output building heights with $\max(S)$
 \end{algorithmic}
\label{alg1}
 \end{algorithm}

\section{Experiments}
\label{sec:Experiments}
To evaluate the proposed method, we compile a dataset consisting of Google Street View images and OSM building footprints. Images are collected by a camera mounted on a moving vehicle. The camera faces the front of the vehicle so that more building rooflines are visible. The image size is $905\times640$. The intrinsic and extrinsic camera parameters are provided, where extrinsic parameters are derived from GPS, wheel encoder, and inertial navigation sensor data \cite{anguelov2010}. We use 400 images covering an area in San Francisco, CA. Map data of the same area is downloaded from OSM, which are in the vector form. We convert the map data into an image plane with a spatial resolution of 0.3 meter. Geo-location information of map data and camera positions is converted into the UTM coordinate system for easy distance computation. Although camera positions in the dataset are more accurate than those solely based on GPS, we observe clear misalignments when projecting map data onto images.  

We apply the proposed method to the dataset with the following parameter setting. For camera position estimation, we use an accumulator corresponding to a local window on the image plane converted from the map. The local window is centered at the initial camera position and of size $40 \times 40$ pixels, which is to correct position errors up to 6 meters. When there are very few detected line support regions corresponding to building edges, estimated camera positions are not reliable. To address this issue, we set an accumulator threshold (set as 1.5) and use estimated camera positions only when the detected peak is above the threshold. We select the largest 5 peaks in an accumulator as candidates of camera positions. To scan an image with projected building footprints for height estimation, we use discrete elevation values from 3 to 100 meters with a step size of 0.2 meter, which cover the range of building heights in the dataset.  

\begin{figure*}
\begin{center}
\includegraphics[width=0.98\textwidth]{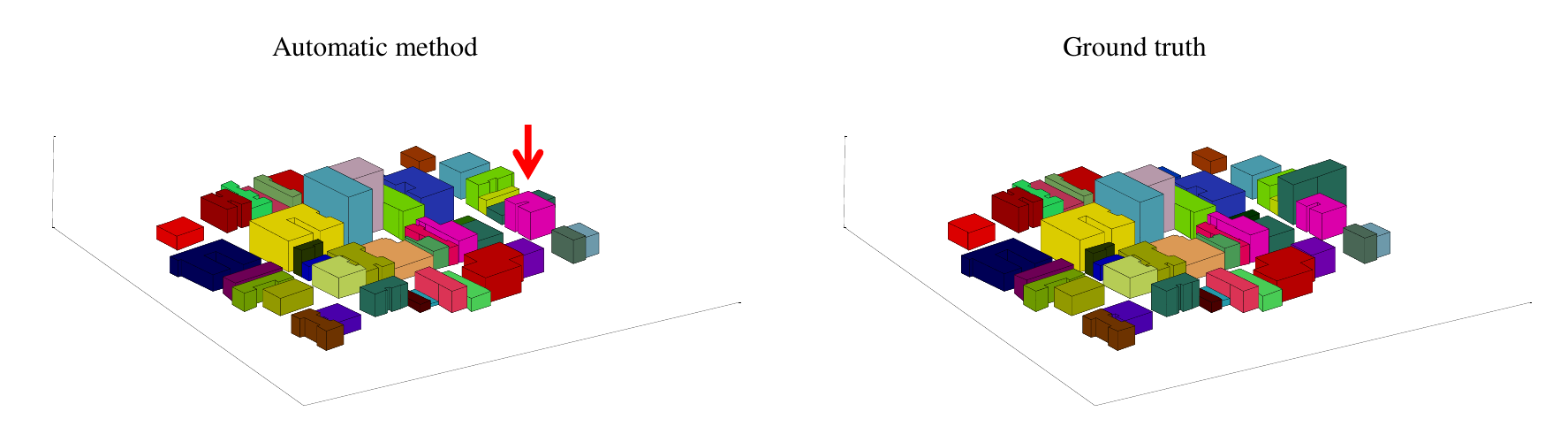}
\end{center}
\caption{3D building models. Left: building models generated by our method. Right: building models using LiDAR derived height information. The red arrow indicates one building with incorrect height estimation.}
\label{fig:bldmdllg}
\end{figure*}

\begin{figure}
\begin{center}
\includegraphics[trim = 0mm 60mm 0mm 20mm, clip, width=0.21\textwidth]{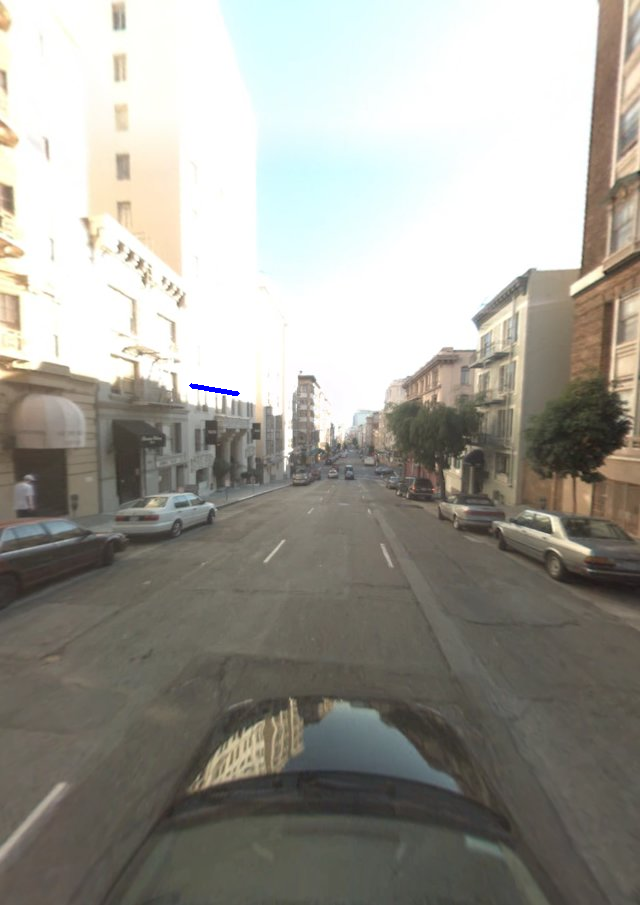}
\includegraphics[trim = 0mm 60mm 0mm 20mm, clip, width=0.21\textwidth]{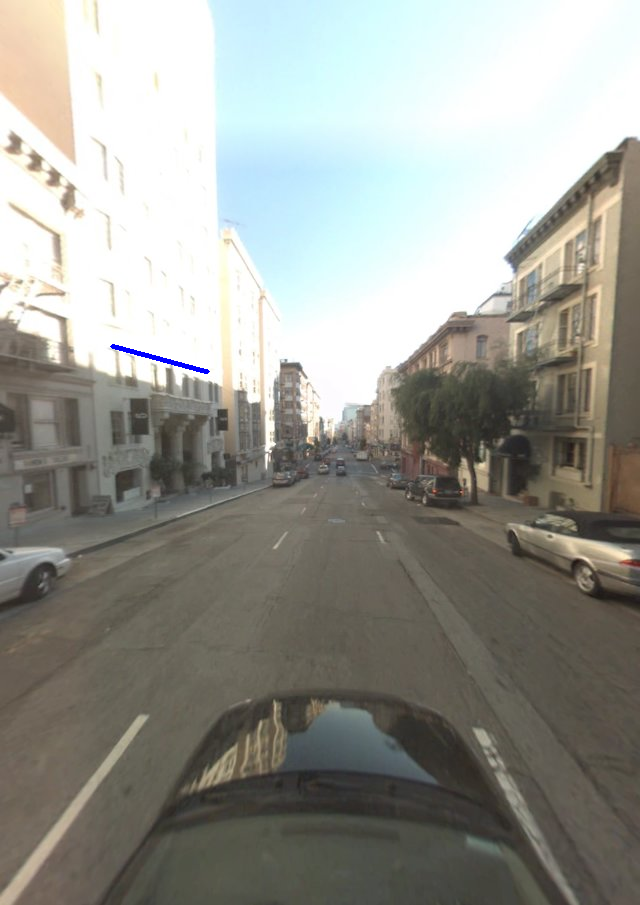}
\end{center}
 \caption{Images causing incorrect height estimation. Blue lines represent detected rooflines. }
\label{fig:err}
\end{figure}

Fig.~\ref{fig:bldmdllg} shows a subset of resulting building models around a city block. We use building height information derived from LiDAR data as ground truth. Building models using ground truth heights are also displayed in Fig.~\ref{fig:bldmdllg}. As we can see, they are very close to each other. There is one building where the estimated height is significantly different from ground truth, which is pointed by a red arrow. We examine the images that the building height is derived from and find that the error is caused by low image quality combined with unusual lighting conditions. Two images are shown in Fig.~\ref{fig:err}, where the proposed method confuses shadow borders as rooflines because the actual rooflines have a extremely low contrast.

To provide quantitative measurements, we calculate errors by comparing the height values from our method with ground truth. We set different error tolerance values (the maximum allowable deviation from ground truth) and compute the percentage of the buildings that have correct height estimation, which is reported in Table~\ref{tab:data1perc}. The results agree with the observation in Fig.~\ref{fig:bldmdllg}. We also evaluate the heights obtained without applying camera position estimation and show accuracy measurements in the table. As can be seen, refined camera positions lead to a clear improvement over initial ones.    

\begin{table}
\small
\begin{center}
\begin{tabular}{c|c c c}
\hline
Error tolerance & 2 m & 3 m & 4 m \\ \hline
Accuracy (refined positions) & 72.2\%  & 83.6\%  & 90.1\%  \\ \hline
Accuracy (initial positions) & 67.3\%  & 78.6\%  & 87.7\%  \\ \hline
\end{tabular}
\end{center}
\caption{Accuracy of height estimation over different error tolerance.}
\label{tab:data1perc}
\end{table}

\begin{figure*}
\begin{center}
\includegraphics[width=0.95\textwidth]{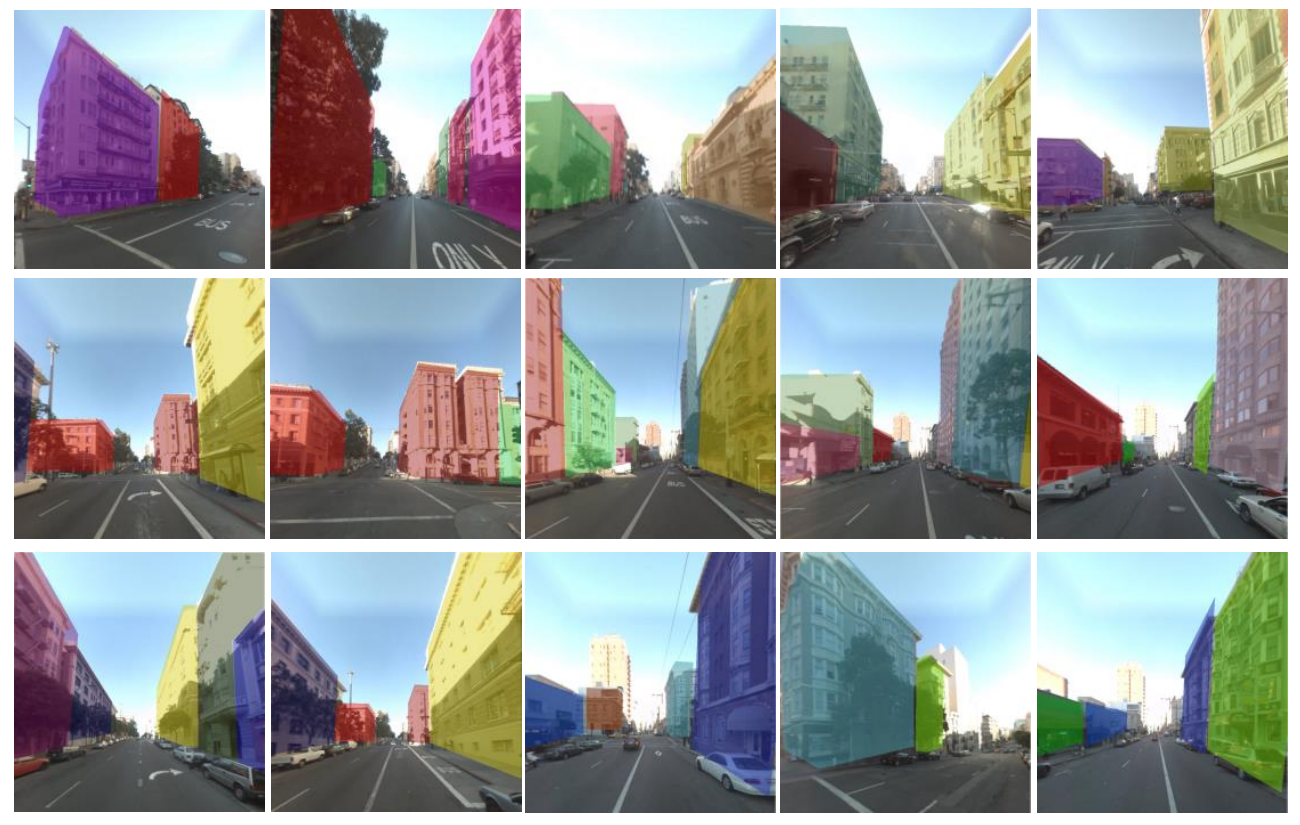}
\end{center}
\caption{Example results of projecting building models onto images. Different buildings are represented by distinct colors.}
\label{fig:bldfcd}
\end{figure*}

\begin{figure}
\begin{center}
\includegraphics[trim = 0mm 90mm 70mm 30mm, clip, width=0.21\textwidth]{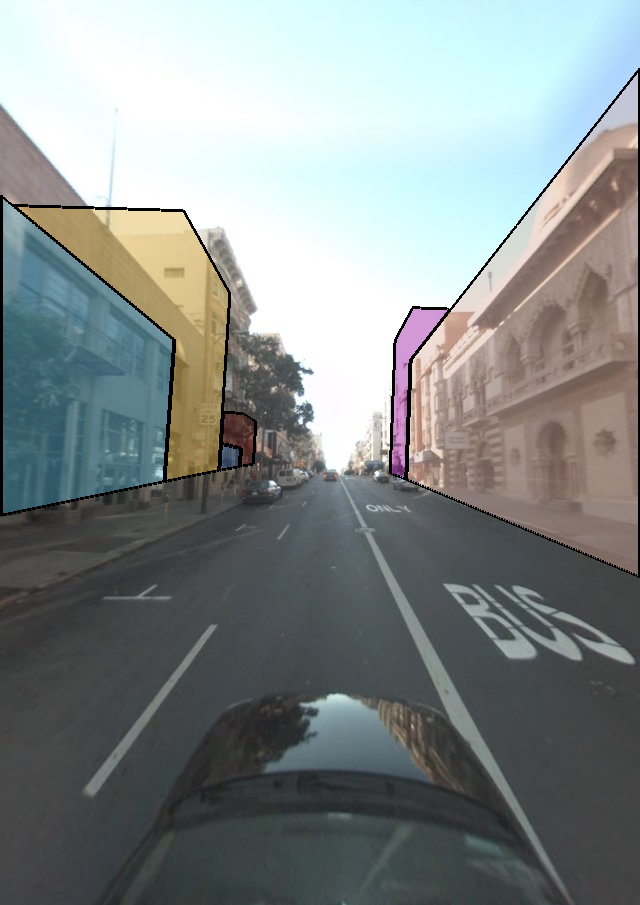}
\hspace{.5mm}
\includegraphics[trim = 0mm 90mm 70mm 30mm, clip,width=0.21\textwidth]{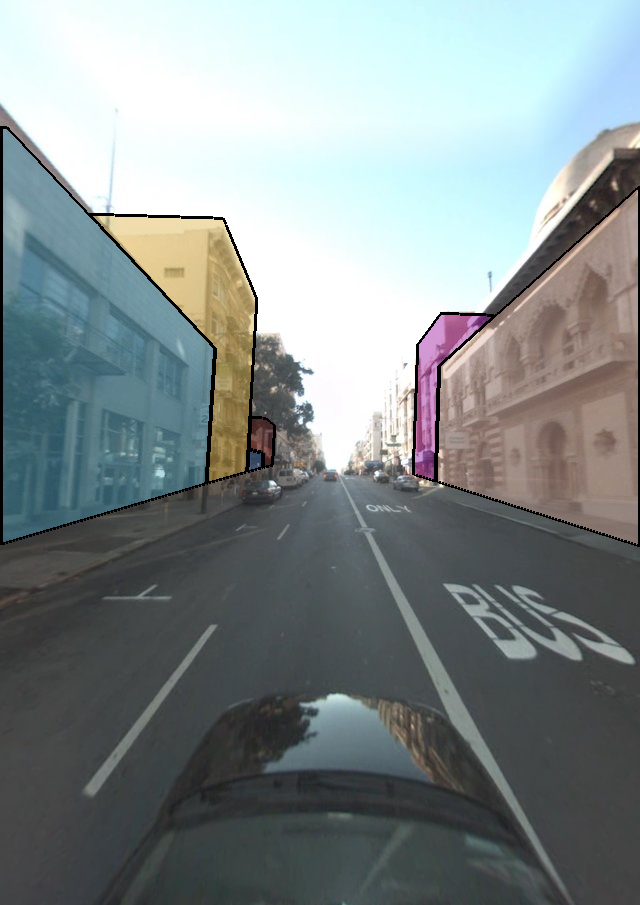}

\end{center}
 \caption{Building facade masks by projecting building models with raw camera position (left) and refined position (right). }
\label{fig:bldseg}
\end{figure}

With images aligned with maps, we label buildings by projecting visible parts of building models onto images. Fig.~\ref{fig:bldfcd} presents example results, where the facade masks obtained by projecting building models are well aligned with building facades in images. To assess the quality of segmented buildings, we select 100 images and manually generate building masks. Each building has a unique label corresponding to the footprint in the map. We calculate an accuracy rate as correctly labeled pixels divided by overall building pixels, where labels of a building that match other buildings are also penalized. Our results reach an accuracy rate of 85.3\%. Camera position accuracy is particularly important for the facade mask quality. We find using raw positions the accuracy drops by 9.2\%. Fig.~\ref{fig:bldseg} shows facade masks from projections using raw and refined camera positions. It can be seen that using the raw position projected buildings severely deviate from the corresponding buildings in images. 

We implement the method using MATLAB on a 3.2-GHz Intel processor. The current version of the code takes on average 3 seconds to process one image, including camera position estimation and height estimation. The efficiency can be further enhanced by using parallel computing resources because each image is processed independently.  

\section{Conclusions}
\label{sec:Conclusions}

We have presented a new approach that fuses 2D maps and street level images, both of which are easily accessible, to perform challenging tasks including building height estimation and building facade identification in images. The method makes effective use of complementary information in data generated from two distinct platforms. Due to the wide availability of input data, the method is highly scalable. Our experiments show that the method performs reliably on real world data.

The proposed method to estimate camera position can work as an add-on for enhancing GPS accuracy. Although in this work we do not measure exact improvements on location accuracy because of the lack of ground truth, we indeed find that even one meter shift of an estimated camera position causes noticeable misalignments between projected building footprints and buildings in images. Worth mentioning is that this method is particularly suited for dense urban neighborhoods, which is often the most challenging situation for acquiring accurate GPS measurements.  

We find that more accurate roofline detection and building edge detection can further improve results. In future work, we will investigate supervised learning based approaches, where a building boundary detector learned from labeled data is expected to provide more meaningful boundary maps.  

\bibliographystyle{abbrv}
\bibliography{sigspatial16}

\end{document}